\documentclass[sigconf]{acmart}

\makeatletter
\def\@ACM@checkaffil{% Only warnings
    \if@ACM@instpresent\else
    \ClassWarningNoLine{\@classname}{No institution present for an affiliation}%
    \fi
    \if@ACM@citypresent\else
    \ClassWarningNoLine{\@classname}{No city present for an affiliation}%
    \fi
    \if@ACM@countrypresent\else
        \ClassWarningNoLine{\@classname}{No country present for an affiliation}%
    \fi
}
\makeatother
\AtBeginDocument{%
  \providecommand\BibTeX{{%
    \normalfont B\kern-0.5em{\scshape i\kern-0.25em b}\kern-0.8em\TeX}}}

\setcopyright{acmcopyright}
\copyrightyear{2018}
\acmYear{2018}
\acmDOI{XXXXXXX.XXXXXXX}

\acmPrice{15.00}
\acmISBN{978-1-4503-XXXX-X/18/06}

\usepackage{multirow} % AADIT'S PACKAGES
\usepackage{booktabs} % AADIT'S PACKAGES
\usepackage{xcolor, soul} % AADIT'S PACKAGES
\sethlcolor{yellow} % AADIT'S PACKAGES
\usepackage{arydshln} % AADIT'S PACKAGES
\usepackage{amsmath} % AADIT'S PACKAGES
\usepackage{algorithm}
\usepackage{algorithmic}
\usepackage{times}
\usepackage{latexsym}
\usepackage{float}

\usepackage[T1]{fontenc}
\usepackage[utf8]{inputenc}

\usepackage{microtype}
\begin{document}

%%
%% The "title" command has an optional parameter,
%% allowing the author to define a "short title" to be used in page headers.
\title{Key-phrase boosted unsupervised summary generation for FinTech organization}

%%
%% The "author" command and its associated commands are used to define
%% the authors and their affiliations.
%% Of note is the shared affiliation of the first two authors, and the
%% "authornote" and "authornotemark" commands
%% used to denote shared contribution to the research.
\author{Aadit Deshpande}
%\authornote{Both authors contributed equally to this research.}
\affiliation{%
  \institution{BITS Pilani, India}
  % \streetaddress{P.O. Box 1212}
  % \city{Dublin}
  % \state{Ohio}
  % \country{USA}
  % \postcode{43017-6221}
}
\email{aadit3003@gmail.com}
%\orcid{1234-5678-9012}
%\author{G.K.M. Tobin}
%\authornotemark[1]
%\email{webmaster@marysville-ohio.com}

\author{Shreya Goyal}
\affiliation{%
  \institution{American Express, India}
 % \streetaddress{1 Th{\o}rv{\"a}ld Circle}
  %\city{Hekla}
  %\country{Iceland}
  }
\email{shreya.goyal@aexp.com}

\author{Prateek Nagwanshi}
\affiliation{%
  \institution{American Express, India}
 % \city{Rocquencourt}
  %\country{France}
}
\email{prateek.nagwanshi@aexp.com}

\author{Avinash Tripathy}
\affiliation{%
 \institution{American Express, India}
 % \streetaddress{Rono-Hills}
 % \city{Doimukh}
 % \state{Arunachal Pradesh}
 % \country{India}
 }
 \email{avinash.tripathy@aexp.com}

% \author{Huifen Chan}
% \affiliation{%
%   \institution{Tsinghua University}
%   \streetaddress{30 Shuangqing Rd}
%   \city{Haidian Qu}
%   \state{Beijing Shi}
%   \country{China}}

% \author{Charles Palmer}
% \affiliation{%
%   \institution{Palmer Research Laboratories}
%   \streetaddress{8600 Datapoint Drive}
%   \city{San Antonio}
%   \state{Texas}
%   \country{USA}
%   \postcode{78229}}
% \email{cpalmer@prl.com}

% \author{John Smith}
% \affiliation{%
%   \institution{The Th{\o}rv{\"a}ld Group}
%   \streetaddress{1 Th{\o}rv{\"a}ld Circle}
%   \city{Hekla}
%   \country{Iceland}}
% \email{jsmith@affiliation.org}

% \author{Julius P. Kumquat}
% \affiliation{%
%   \institution{The Kumquat Consortium}
%   \city{New York}
%   \country{USA}}
% \email{jpkumquat@consortium.net}

%%
%% By default, the full list of authors will be used in the page
%% headers. Often, this list is too long, and will overlap
%% other information printed in the page headers. This command allows
%% the author to define a more concise list
%% of authors' names for this purpose.
\renewcommand{\shortauthors}{Aadit Deshpande, et al.}

%%
%% The abstract is a short summary of the work to be presented in the
%% article.
\begin{abstract}
 With the recent advances in social media, the use of NLP techniques in social media data analysis has become an emerging research direction. Business organizations can particularly benefit from such an analysis of social media discourse, providing an external perspective on consumer behavior. Some of the NLP applications such as intent detection, sentiment classification, text summarization  can help FinTech organizations to utilize the social media language data to find useful external insights and can be further utilized for downstream NLP tasks.  Particularly, a summary which highlights the intents and sentiments of the users can be very useful for these organizations to get an external perspective. This external perspective can help organizations to better manage their products, offers, promotional campaigns, etc. However, certain challenges, such as a lack of labeled domain-specific datasets impede further exploration of these tasks in the FinTech domain. To overcome these challenges, we design an unsupervised phrase-based summary generation from social media data, using  \textit{'Action-Object'} pairs (intent phrases). We evaluated the proposed method with other key-phrase based summary generation methods in the direction of contextual information of various Reddit discussion threads, available in the different summaries. We introduce certain “Context Metrics” such as the number of \textit{Unique words}, \textit{Action-Object pairs}, and \textit{Noun chunks} to evaluate the contextual information retrieved from the source text in these phrase-based summaries. We demonstrate that our methods significantly outperform the baseline on these metrics, thus providing a qualitative and quantitative measure of their efficacy. Proposed framework has been leveraged as a web utility portal hosted within Amex.
 
\end{abstract}

%%
%% The code below is generated by the tool at http://dl.acm.org/ccs.cfm.
%% Please copy and paste the code instead of the example below.
%%


%%
%% Keywords. The author(s) should pick words that accurately describe
%% the work being presented. Separate the keywords with commas.

%% A "teaser" image appears between the author and affiliation
%% information and the body of the document, and typically spans the
%% page.
% \begin{teaserfigure}
%   \includegraphics[width=\textwidth]{sampleteaser}
%   \caption{Seattle Mariners at Spring Training, 2010.}
%   \Description{Enjoying the baseball game from the third-base
%   seats. Ichiro Suzuki preparing to bat.}
%   \label{fig:teaser}
% \end{teaserfigure}

% \received{20 February 2007}
% \received[revised]{12 March 2009}
% \received[accepted]{5 June 2009}

%%
%% This command processes the author and affiliation and title
%% information and builds the first part of the formatted document.
\maketitle

\section{Introduction}

% ********FIG 1: What we are doing - Overview********
\begin{figure}[!t]
\small
\centering
\begin{tabular}{p{0.9\linewidth}}
\toprule
% \multicolumn{1}{c}{\textbf{Input} (Thread)}                                                      \midrule                                           
\textbf{Post}: \\ 
Bank Rewards Checking Account.  \\  \midrule                                                                              
\textbf{Comments}: I hope they add the ability to make it a \hl{joint account with a spouse}. \textbf{\textit{\textless sep\textgreater}} Nice to see they're attempting to \hl{compete} with newer \hl{online banks}. \textbf{\textit{\textless sep\textgreater}} When they first \hl{opened} up their \hl{HYSA} they eventually had a \hl{signup bonus}. I wonder if they’ll do something like that at some point \textbf{\textit{\textless sep\textgreater}} The main difference between this and the HYSA is that you’ll be able to makes charges and \hl{get rewards} for purchases \textbf{\textit{\textless sep\textgreater}} If you link your new Checking account to your credit card there’s no way you would \hl{get MR points} on that transaction, right? \textbf{\textit{\textless sep\textgreater}} Can you \hl{use Zelle} through the account?

\\ \midrule
\multicolumn{1}{c}{{\textbf{Phrase-based summaries}}}                                                                                                  \\ \midrule
%\textbf{Verb-Noun intents}: 
\textcolor{blue}{signup bonus} ; 
\textcolor{blue}{compete online banks} ; 
\textcolor{blue}{opened HYSA} ; 
% \textcolor{blue}{get rewards} ; 
\textcolor{blue}{get MR points} ;         
\textcolor{blue}{use Zelle}                                           \\
% \textbf{Positive Aspects}: 
% %international ATM fees ; 
% joint account ; 
% HYSA signup bonus ; 
% opened HYSA ; 
% make charges ;
% get rewards ; 

\\ \bottomrule
\end{tabular}
\caption{An illustration of phrase-based summary generation for a Reddit Thread (post and its comments separated by the token \textbf{\textit{\textless sep\textgreater}}).
}
\label{fig1}
\end{figure}

% Para 1 Background, Broad

With the advent of social media, the analysis of data available on these platforms has the potential to provide a gold mine of actionable insights to business organizations. With the ubiquity of social media, the amount of available data has skyrocketed, and social media reflects popular discourse and discussion in a larger societal context. Thus, building a social media knowledge ecosystem in addition to existing data analysis platforms can help an organization to better understand its consumers’ preferences and reactions to the services they offer. These insights help businesses form an understanding of the ``external perspective”, directly from the source - the consumers. This additional analysis can have long-term positive impacts and reveal insights that may not be surfaced through traditional channels like customer service call transcripts. Summarizing the discussions available on these social media platforms can give a precise external perspective to an organization about their products, offers, competitors etc, while helping them to make decision for their future products and offers. Hence, in this study, we focus our attention on the problem of phrase based unsupervised summary generation from the Reddit discussions data, with the goal of developing a broader understanding of social media discussions related to financial organization's products and services. 

% With the recent advancements in social media platforms, the role of analysing the data available on these websites about a certain organization or a group of individuals has become very critical. Social media platforms have become an epitome for modern day structure of the society and discussions flowing on them can bring out the ``Talk of the Town" in a specific context. Social media data analysis if done in a right way, may lead to some positive and long term impacts on the business of an organization. Building a social media knowledge ecosystem in addition to the existing data analysis platforms can help an organization to better understand the user preferences and their reactions to a certain product or service. Some of these insights may not come out using the traditional approaches of using call transcripts data or feedback systems. Hence, to have a broader understanding of social media communications and discussions related to American Express products and services, an intent based insight generation scheme is proposed in this paper. The proposed method takes thousands of Reddit posts and comments data into account and tries to generate phrase based insights which describes the underlying context of these discussions groups.  

% Para 2 - Difficulty
In the last decade, NLP techniques have been successfully applied to social media data in the FinTech context to improve business operations and understand customer opinions. They have been applied to problems like sentiment analysis, stock value prediction, document summarization, intent detection, and so on. A facet that has received particular attention is sentiment analysis from social media data and its utility in predicting trends in the stock market and share prices \cite{day2016deep,aji2019sentiment,karalevicius2018using, esichaikul2018sentiment,sangsavate2019stock}. 
%\textcolor{red}{Write some literature about summarization and connect with intent detection}.
Lack of publicly available data and the cost of generating gold standard labels has been a challenge for multiple NLP tasks such as summarization, intent detection, which has lead to the requirement of more unsupervised or semi supervised approaches. Some of the work in this direction are, \cite{abdaljalil2021exploration,la2020end} for summarizing financial reports and FinTech data, \cite{weld2022survey} for task of identifying the user’s intention. Despite availability of some of the labelled datasets for summarization, intent detection, its difficult to solve domain specific problems with them.
% Summarising the financial reports and FinTech data is another explored problem in literature \cite{abdaljalil2021exploration,la2020end}, where the training dataset is labelled and contains gold standard summaries of large financial reports.
%Generating summaries from these financial reports in an unsupervised manner still remains a challenge due to lack of domain specific language models for financial textual data. Also, the cost of generating gold standard summaries for these reports is another factor which makes this task challenging. 
% Another important but not well-explored application in FinTech involves intent detection, the task of identifying the user’s intention \cite{weld2022survey}.
% Certain factors such as lack of publicly available domain specific data, cost of obtaining labels for the new/unseen data, make intent detection a challenging area of study. 
%First, the lack of publicly available domain-specific data makes training models for FinTech social media data difficult.
Most studies on intent detection focus on the spoken language understanding (SLU) datasets - ATIS  \cite{hemphill-etal-1990-atis}, and SNIPS \cite{https://doi.org/10.48550/arxiv.1805.10190}, which were originally designed for mobile voice assistants, and hence do not generalize well. Secondly, the current formulation of intent detection frames it as a multi-class classification problem. Though pretrained models perform well \cite{https://doi.org/10.48550/arxiv.2004.14848}, the intent class labels designed for ATIS and SNIPS do not apply to other domains and are too restrictive for real-world data. Finally, this supervised definition of intent detection means crowd-sourcing is the only viable option for creating domain-specific social media datasets. Crowd-sourcing comes with its own set of problems, as it is often expensive and presents problems with annotator bias \cite{eickhoff2018cognitive}. Unsupervised methods in intent detection have been explored in the past \cite{popov2019unsupervised, dopierre2021protaugment, siddique2021generalized}, but they do not pertain to FinTech data. Traditionally employed in SLU, intent detection has great potential in this domain. Not only can it summarize and simplify lengthy social media discussions, but these salient features can also form representative features for various downstream NLP tasks. 
% Para 3 In this paper we...

In this regard, we propose a completely unsupervised method for phrase-based summary generation, specifically for FinTech social media data. We build a database of posts and comments (threads) from the social news platform Reddit and extract key intent phrases to summarize lengthy threads. None of our data is annotated with summary text or keywords, and we propose a flexible unsupervised approach.
In this work the summary of the Reddit discussions is represented by intent keywords in the data. For intent phrase extraction, we follow a two-stage pipeline by first identifying different categories of 'Action-Object' pairs (tokens matching pre-defined rules), followed by a scoring function to rank them. Additionally for comparison purpose,
we perform aspect-based sentiment analysis and identify the key phrases that are associated with the most positive or negative comments.
%Additionally we compared the generated intent keyword boosted summary with the aspect-sentiment keywords based summary.  
Figure \ref{fig1}, demonstrates an overview of our pipeline by identifying the most salient intent phrases from a Reddit thread. These intent phrases are further used as a as vector representations to cluster the post corpus and generate cluster-level summaries to gather insights from large corpus of customer feedback on Reddit.  We compared our method by generating summary using an existing keyphrase extraction method, ‘Yake’ \cite{campos2018yake} and  aspect-based sentiment phrases. For evaluation, we additionally introduce ‘context metrics’ to evaluate the contextual information in the respective cluster summaries. 

% Para 4 Contributions
The main contributions of this paper are as follows. (1) We propose a new unsupervised method for phrase-based summary generation using intent key phrases. (2) We study publicly available social media data (pertaining to FinTech), (without crowdsourcing annotation). Finally, (3) we suggest new metrics to evaluate the contextual information retrieved in the clusters and find that our method outperforms existing keyphrase based summary generation methods. Thus, our goal in this paper is to demonstrate that intent based summary can generate meaningful insights in domain-specific financial text without supervision or labeled datasets.
In Section \ref{sec:methodology}, data collection and description of the end-to-end summary generation pipeline is described. Section \ref{sec:mexperiments} experimental setup, qualitative and qualitative results are demonstrated.
Section \ref{sec:conclusion} conclude the paper and briefly discuss the limitations of the study and future directions.

% . We then proceed to an in-depth description of the end-to-end intent detection pipeline. In section 3, we discuss the experimental setup, as well as the evaluation cluster metrics, and discuss the results of our method compared to existing baselines. Next, in section 4, we present some qualitative examples of our intent detection process and discuss the results in more detail. Finally, in section 5, we conclude the paper and briefly discuss the limitations of the study and future directions.

% ********TABLE 1: Dataset Stats********
\begin{table}[!t]
\small
\centering

\begin{tabular}{lc}
\midrule
No. of Posts                   & $5370$ \\
No. of Comments                & $33823$ \\ \midrule
Avg. No. of Comments Per Post  & $6.29$  \\
Avg. Post length (in words)    & $81.32$ \\
Avg. Comment Length (in words) & $36.22$ \\
Avg. Thread Span               & $3.37$  \\ \midrule
\end{tabular}
%}
\caption{Dataset statistics for our Reddit Post and Comments Data. Thread Span refers to the time between the first and most recent comment on the thread (in days). }
\label{table1}
\end{table}

% ***************************METHODS***************************
\section{Methodology} 
\label{sec:methodology}
\subsection{Data Collection} 
For social media insight generation, we consider the news aggregation social network – 'Reddit'. There are a few factors that motivate this choice. Firstly, Reddit has an active user base ($\sim450$ million monthly users), and a large amount of data is readily available. Next, the nature of the Reddit platform lends itself to discourse-like natural language, where comments may answer, query, or refute the original posts. Finally, Reddit communities are actively moderated by user admins, who ensure that content posted by users complies with community rules regarding quality and relevance. These qualities make Reddit data especially pertinent to tasks like intent extraction and summarization.

% ********FIG 2: Intent Detection Pipeline********
\begin{figure*}[!t]
    \centering
    \includegraphics[width = \textwidth]{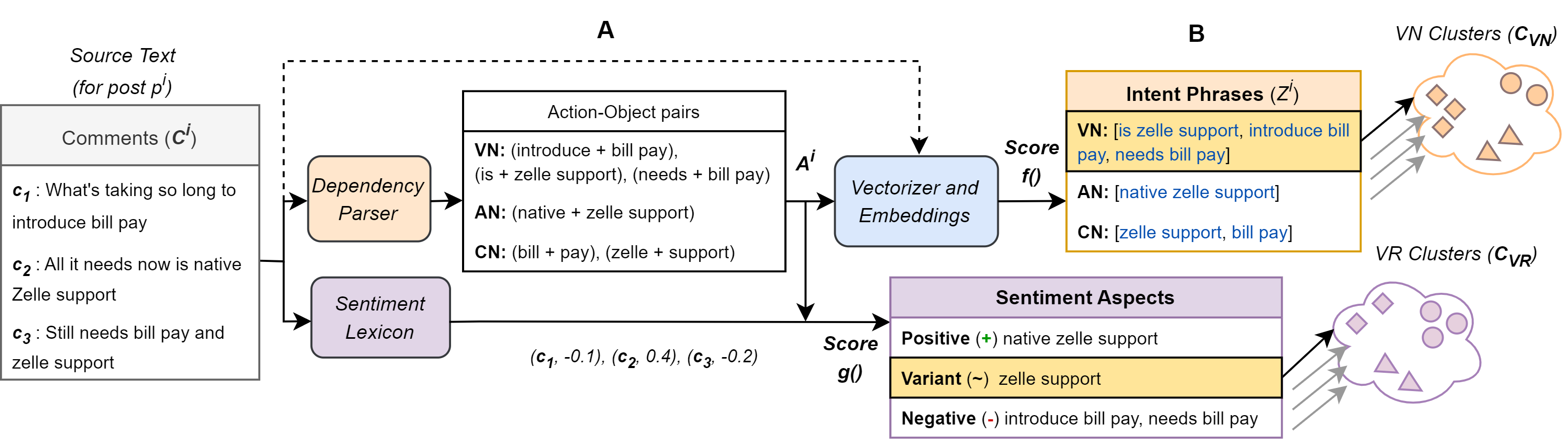}
    \caption{The intent detection framework. Action-Object pairs, $A^i$ (Verb-Noun, Adj-Noun and Compound Noun) are generated from the comments under a thread ($C^i$) using a dependency parser. The action-object pairs are then scored using their tf-idf embedding score according to the function $f()$ to obtain the Intent phrases ($Z^i$). Additionally, the comments are assigned sentiment scores, and the respective action-object pairs (aspects) are scored using the sentiment scoring function $g()$.}
    \label{fig2}
\end{figure*}

Reddit is organized into communities based on user interests known as subreddits (prefixed with “r/” ).  In this paper, we refer to a post and its top-level comments as a ‘thread,’ and we focus on the comments for summary generation and intent analysis. We create a novel data scraper using the python wrappers for the Reddit API (PRAW and PSAW) to collect a database of Reddit threads (34K unique comment entries and $\sim 5.4$K unique posts) over 10 months. We pre-process the raw data by deleting comments under posts tagged ‘Low-Quality Post,’ ($\sim20$\% of the original data) and deleting comments from deleted accounts or bots. Reddit comments under a thread are organized in a tree-like hierarchy and we only consider top-level comments for our analysis.  Table \ref{table1} shows the dataset statistics.

% % ********EQ 0: Net Interaction Sentiment********
% \begin{equation}
%     NIS(p^i) = \frac{1}{|C^i|} \sum_{c \in C^i} SS_c
%     \label{equation0}
% \end{equation}

\subsection{Problem Definition}
Let $D_r$ be the Reddit posts and comments database, which is a set of tuples $(p^i, C^i)$ (threads), where $p^i$ represents a post and $C^i = \{c_1^i, c_2^i, ..., c_{|C^i|}^i \}$, represents the respective comments. We define intent keyphrase detection as the process of generating the sequence $Z^i = \{z_1^i, z_2^i, ..., z_m^i\} $ of the top $m$ key phrases from the text $C^i$, without supervision. We refer to each phrase $z_j^i$, as an intent phrase, and we extract the top ten key intent phrases $(m = 10)$ from each thread in $D_r$. Hence, $Z^i$ is the sequence of phrases summarizing the discussion appearing in the comments for these posts. In this paper we define $Z^i$ as the phrase based thread summary. 

The first stage in our intent detection pipeline produces a sequence of action-object pairs within a thread, $A^i = \{(a_1^i, o_1^i), (a_2^i, o_2^i),  ...\} $. Inspired by previous work on unsupervised intent detection \cite{liu2021open}, we define intent phrases as \textit{Action-Object pairs}, where Actions represent user actions, tasks, or commands, and Objects represent entities that the actions may operate on. In the next stage, the action-object pairs in $A^i$ are scored and ranked to generate the sequence of intent features $Z^i$. Each $Z^i$ is then used as a set of intent features to cluster the threads using hierarchical agglomerative clustering to generate cluster-based summaries $C = \{c_1, c_2, ..., c_N\}$. More details regarding the Action-Object pairs, intent detection pipeline, and clustering methods are presented in the following sections.

\subsection{Intent Detection Pipeline}
We follow a two-stage pipeline for unsupervised intent detection of the Reddit threads, as shown in Fig. \ref{fig2}. Stage (A) involves the identification of Action-Object pairs using dependency parsing and proximity rules. This is followed by scoring and ranking the intent phrases to generate a phrase-based summary at the thread level in stage. Scoring of intent phrases is performed, to pick the top $n$ Action-Object pairs out of multiple extracted pairs. 
(B). Finally, we use the observation-level (thread) summaries to cluster the posts and generate group-level summaries. 

\subsubsection{Action-Object pair generation (A):}
First, we label tokens in the comments, $C^i$ (for the post $p^i$), using the spaCy dependency parser and Part-of-speech tagger \cite{honnibal2015improved}. These labels are then used to generate ‘Action-Object’ pairs according to a series of predefined grammar rules based on prior work on intent discovery through dependency parsing \cite{liu2021open}. The grammar rules and part-of-speech labels we consider as ‘Action-Object’ pairs are as follows: Verb-Noun (VN), Adjective-Noun (AN), Compound-Noun (CN), Verb-Object (VO), and Adjectival Complement-Prepositional Object (AP).

If matching tokens are within three words of each other, the pair is added to its respective category. We also consider multiple-word compound nouns as ‘Objects’ and distinguish between Verb-Noun and Verb-Object phrases. For additional insight, we introduce a sixth rule for ‘Negation,’ which involves any action-object pairs negated in the text. This process is repeated for the comments associated with every post to generate the six kinds of action-object phrases ($A^i$) mentioned above.

% ********TABLE 2: Comparison of VN, VR, Yake on Cluster + Context metrics********
\begin{table*}[!t]
\small
\centering
\begin{tabular}{ccccccc}
\cmidrule{1-7}
\multirow{2}{*}{Method}        & \multicolumn{3}{c}{Context metrics (per cluster)}                                                          & \multicolumn{3}{c}{Cluster metrics}              \\ \cmidrule{2-7}
                               & Unique Words           & Action-Object Pairs    & \multicolumn{1}{c}{Noun Chunks}            & Silhouette     & CH             & DBI            \\ \cmidrule{1-7}
\multicolumn{7}{c}{\textit{50 Clusters}}                                                                                                                                         \\ \cmidrule{1-7}
Yake                           & \textbf{589.1 ± 128.8} & 266.6 ± 88.1           & \multicolumn{1}{c}{861.7 ± 383.7}          & \textbf{0.030} & \textbf{13.41} & \textbf{5.97} \\
Variant Sentiment Aspects (VR) & 473.7 ± 106.5          & 344.9 ± 123.7          & \multicolumn{1}{c}{631.5 ± 233.3}          & 0.001          & 7.09           & 8.19           \\
Verb Noun Intents (VN)         & 588.8 ± 119.9          & \textbf{433.8 ± 142.5} & \multicolumn{1}{c}{\textbf{564.2 ± 232.9}} & 0.020          & 10.36          & 6.54            \\ \cmidrule{1-7}
\multicolumn{7}{c}{\textit{100 Clusters}}                                                                                                                                          \\ \cmidrule{1-7}
Yake                           & 331.4 ± 87.6           & 133.2 ±51.6            & \multicolumn{1}{c}{431.1 ± 222.7}          & \textbf{0.030} & \textbf{9.79}  & \textbf{5.03}  \\
Variant Sentiment Aspects (VR) & 280.1 ± 71.5           & 172.7 ± 72.9           & \multicolumn{1}{c}{315.2 ± 138.9}          & -0.006         & 4.94           & 6.73          \\
Verb Noun Intents (VN)         & \textbf{347.8 ± 86.9}  & \textbf{217.36 ± 85.4} & \multicolumn{1}{c}{\textbf{281.4 ± 126.3}} & 0.020          & 7.38           & 5.47         \\ \cmidrule{1-7}
\end{tabular}
    \caption{Clustering performance for 50 and 100 clusters, of an existing keyphrase extraction method (Yake) and our methods. VN and VR achieve comparable performance to Yake on various clustering evaluation metrics. However, VN and VR yield more unique words and action-object pairs (per cluster), providing more context information than noun chunks. The best results (Mean $\pm$ SD) are boldfaced.}
\label{table2}
\end{table*}

\subsubsection{Intent Phrase scoring (B):}
In the next stage, we score and rank these action-object pairs in $A^i$ (intent phrases) using the composite tf–idf scores of their constituent words (The tf-idf vectorizer is trained on the organization's internal complaints data). The sum of the tf-idf scores  over the document axis yields a 1-D matrix of the composite tf-idf scores for each word in the vocabulary ($\sim$97K unique words). Tf-idf embeddings are chosen for their efficiency and the fact that the embedding size matches exactly with the vocabulary size. 
The Intent Scoring Function is defined as $f(z_x^i) = \frac{\sum\limits_{w \in (a_x^i, o_x^i)} TI_w}{|(a_x^i, o_x^i)|}$. Here,
$f()$ represents the intent scoring function which assigns the score to the intent phrase $z_x^i$, by averaging the tf-idf scores, $TI$ for each word $w$. Out-of-vocabulary words at the test time are assigned a score of zero. We use the top ten highest-scoring intent phrases in each rule to determine the most salient intent phrases for a thread. Thus, we obtain a list of the most relevant intent phrases, $Z^i$ (for each rule), and the intent phrase-based summary for each thread. 

% ********EQ 1: Intent Scoring function********
% \begin{equation}
%     f(z_x^i) = \frac{\sum\limits_{w \in (a_x^i, o_x^i)} TI_w}{|(a_x^i, o_x^i)|}
%     \label{equation1}
% \end{equation}

The intent phrases obtained in the steps discussed above are utilized as features for generating phrase based summaries of the lengthy discussions in Reddit data. The detailed process is described in the coming section. 

\subsubsection{Intent-based Summaries:}
We proceed to use the observation-level summaries as vector representations to cluster the Reddit threads.  We perform agglomerative clustering using the CLUTO clustering toolkit's 'vcluster functionality \cite{karypis2002cluto} to generate cluster-level summaries $C_{VN} = \{c_{VN_1}, c_{VN_2}, ..., c_{VN_n}\}$ and related statistics. We use the top ten intent phrases from the rule ‘Verb-Noun’ (VN, ranked by the intent scoring function $f()$) as the vector representations of each post for agglomerative clustering.

\subsubsection{Sentiment key-phrase based summaries:}
Additionally, we perform phrase-based sentiment analysis using the intent phrases, $A^i$. We consider the intent phrases to be various ‘aspects’ of the thread and check their occurrence in the comments $C^i$. We calculate the sentiment scores of the comments and average them to obtain the sentiment score for each occurring ‘aspect.' Aspect sentiment scoring function is defined as $g(z_x^i) = \frac{\sum\limits_{t \in \{ t \in C^i | z_x^i \in t\}} SS_t}{|{\{t \in C^i | z_x^i \in t\}}|}$.
Here, $g()$ represents the aspect sentiment scoring function which assigns the score to the intent phrase $z_x^i$, as the average of sentiment scores $SS_t$ for each comment $t$ that contains an incidence. In addition, we use the top ten phrases from the ‘Variant Sentiment aspects’ (VR, ranked by the scoring function $g()$) for agglomerative clustering of the threads and generate another set of cluster-level summaries, $C_{VR} = \{c_{VR_1}, c_{VR_2}, ..., c_{VR_n}\}$. 

% ********EQ 2: Sentiment Scoring function********
% \begin{equation}
%     g(z_x^i) = \frac{\sum\limits_{t \in \{ t \in C^i | z_x^i \in t\}} SS_t}{|{\{t \in C^i | z_x^i \in t\}}|}
%     \label{equation2}
% \end{equation}

We repeat this process for all the intent phrases and then use these scores to produce three new rankings. Positive Phrases - Intent phrases with the highest sentiment scores, thus, generally associated with positive or neutral comments. Variant Phrases - Intent phrases with the most variance in their sentiment scores (associated with comments representing mixed opinions). Negative Phrases - Intent phrases with the lowest sentiment scores, thus, generally associated with mostly negative comments.
% \begin{itemize}
% \item Positive Phrases - Intent phrases with the highest sentiment scores, thus, generally associated with positive or neutral comments.
% \item Variant Phrases - Intent phrases with the most variance in their sentiment scores (associated with comments representing mixed opinions).
% \item Negative Phrases - Intent phrases with the lowest sentiment scores, thus, generally associated with mostly negative comments.
% \end{itemize}

Fig. \ref{fig2} demonstrates the end-to-end intent detection framework, starting from the comments in a thread to stage Action-Object pair generation, and Intent phrase scoring, culminating in the cluster-level summaries. It also shows the phrase-based sentiment analysis, similarly resulting in group-level (cluster) summaries.

% ********FIG 3: Graphs for 50, 100 Clusters -> Context metrics********
\begin{figure*}
    \centering
    \includegraphics[width = \textwidth]{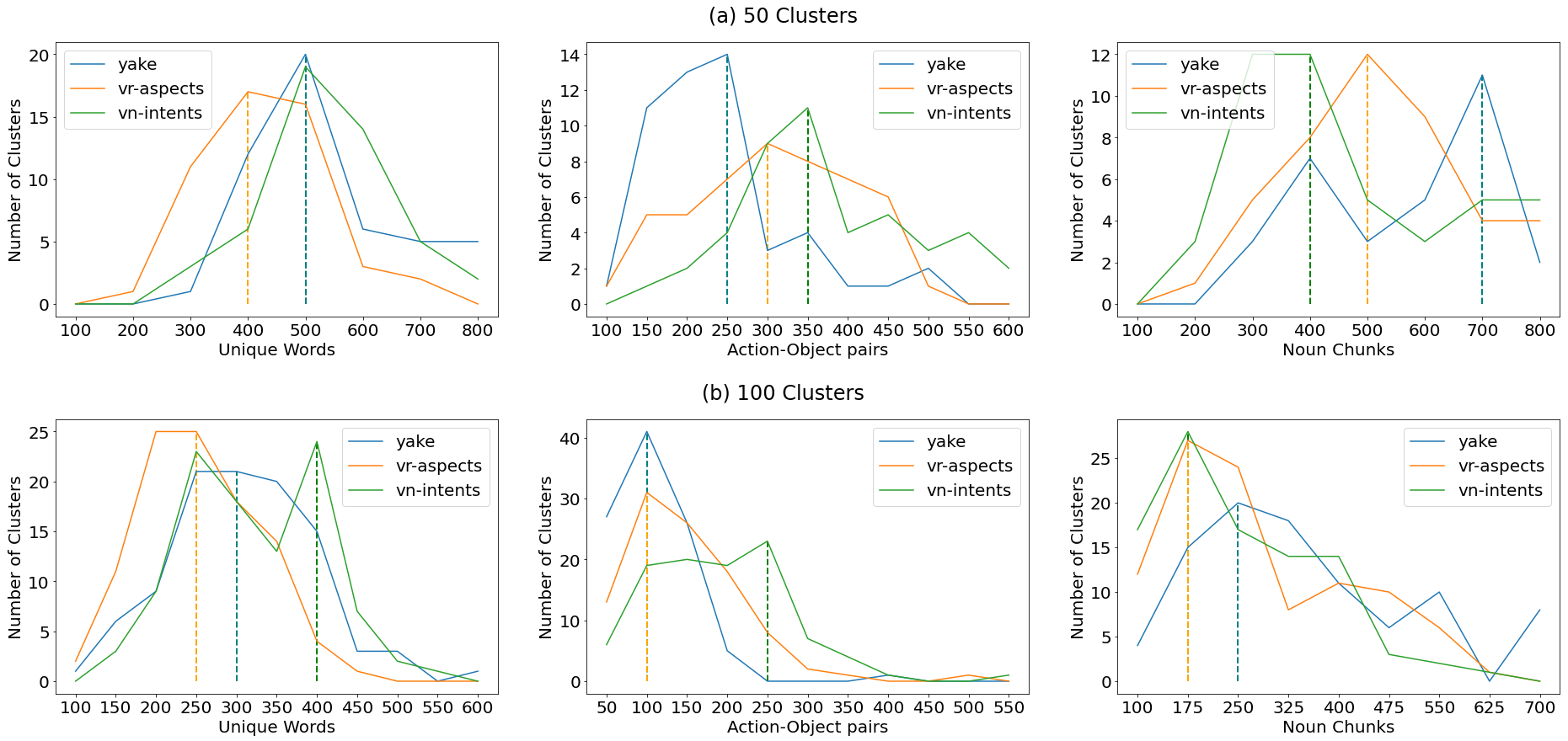}
    \caption{(a) shows the histograms for the context metrics (per cluster) with each method for 50 clusters, and similarly, (b) demonstrates these for 100 clusters. VN and VR outperform Yake in Action-Object pairs and Unique words. This trend is reversed in the case of Noun Chunks, where Yake retrieves the most, indicating that its keyphrases contain less context information. The dotted lines represent the peak values for each method.}
    \label{fig3}
\end{figure*}

% ***************************EXPERIMENTS***************************
\section{Experiments and Results}
\label{sec:mexperiments}

\subsection{Experimental Setup}
We use the Reddit dataset described in the previous section to evaluate our intent-phrase generation pipeline. Since we have no labels to test our results against, we use these phrase-based summaries to cluster the dataset of $5370$ Reddit posts (see Table \ref{table1}) as mentioned above. Using CLUTO’s ‘vcluster’ functionality, we perform agglomerative clustering using vector representations of the data points (posts) and generate statistics regarding the quality of the clusters. We use these statistics to judge the intent phrase summaries of our methods. We use the ‘Yake’ unsupervised keyword extraction system \cite{campos2018yake} as the baseline to compare with key phrases extracted via - VN and VR. For Yake, we similarly use the top ten n-grams (n= 2, 3) and set the deduplication threshold to $0.9$. We tried a number of values for the clustering parameter ($NClusters = 20, 30, 50, 100)$ and, based on the subjective quality of the clusters, chose to continue our analysis with $50$ and $100$ clusters.

\subsection{Evaluation Metrics}
We evaluate the quality of clustering with key-phrase feature using Silhouette Score \cite{rousseeuw1987silhouettes}, Calinski-Harabasz Index (CH) \cite{calinski1974dendrite}, Davies-Bouldin Index (DBI) \cite{davies1979cluster}. Although, in the context of summary generation from social media discussion for an organization, more than the quality of clustering, it is important to evaluate the quality of the summary with respect to the contextual information carried by that cluster summary. In this regard, we introduce another set of cluster-level metrics with the intention of evaluating the context information in the resultant clusters $C = \{c_1, c_2, ..., c_N\}$. Each $c_i$ is a sequence of the phrase-based summaries for the respective posts in that cluster:
% The agglomerative clustering using the phrase features is evaluated using  Silhouette Score \cite{rousseeuw1987silhouettes}, Calinski-Harabasz Index (CH) \cite{calinski1974dendrite}, Davies-Bouldin Index (DBI) \cite{davies1979cluster}. Though the above metrics provide an idea of the clustering, they do not provide much information regarding the quality of the phrase-based summaries. 

\begin{itemize}
\item Unique Words (per cluster):  Unique Words (per cluster) are calculated as $u(C) = \frac{1}{N} \sum_{i = 1}^{N} |\{c_i\}|$. We define this metric as the average cardinality of $c_i$ over $N$ clusters (the average number of unique words per cluster). We assume that word repetition in the phrase vectors is a poor indicator of context information and aspects of the source text, during keyphrase extraction. Thus, higher values (more unique words per cluster) indicate better context retrieval in the phrase-based summaries. 

% ********EQ 3: Unique words definition********
% \begin{equation}
%     u(C) = \frac{1}{N} \sum_{i = 1}^{N} |\{c_i\}|
%     \label{equation3}
% \end{equation}

\item Noun Chunks (per cluster): Noun Chunks (per cluster) are calculated as $ n(C) = \frac{1}{N} \sum_{i = 1}^{N} | \{t_1, t_2 \in NN \mid t_1, t_2 \in c_i,    dist(t_1, t_2) = 1 \}|$. We calculate this value as average number of consecutive Noun-Noun pairs in each sequence $c_i$. Though more noun chunks (phrases consisting of only nouns) in phrase-based vectors provide information about entities in the source text, they do not provide any contextual information. 

% ********EQ 4: Noun Chunks definition********
% \begin{equation}
% \begin{split}
%    n(C) = \frac{1}{N} \sum_{i = 1}^{N} | \{t_1, t_2 \in NN \mid t_1, t_2 \in c_i, \\   dist(t_1, t_2) = 1 \}|
%     \label{equation4}   
% \end{split}
% \end{equation}

\item Action-Object pairs (per cluster): We calculate this value as the average number of Action-Object pairs (Adj-Noun, Verb-Noun) within 3 words in each sequence $c_i$. Unlike noun chunks, more instances of Action-Object pairs are far richer sources of context information, like intensity, location, and subject-object relations. $a(C) = \frac{1}{N} \sum_{i = 1}^{N} | \{t_1, t_2 \in AO \mid t_1, t_2 \in c_i,   dist(t_1, t_2) \leq 3 \}|$
Thus, a higher value of Action-Object pairs, in conjunction with a lower value of noun chunks, indicates better context retrieval.

% ********EQ 5: AO Pairs definition********
% \begin{equation}
% \begin{split}
%     a(C) = \frac{1}{N} \sum_{i = 1}^{N} | \{t_1, t_2 \in AO \mid t_1, t_2 \in c_i, \\   dist(t_1, t_2) \leq 3 \}|
%     \label{equation5}   
% \end{split}
% \end{equation}
 \end{itemize}

% ***************************RESULTS***************************
\subsection{Results}
\label{sec:results}

Table \ref{table2} shows the results of clustering using the three phrase-based vector representations- Yake, Verb-Noun Intents (VN), and Variant Sentiment aspects (VR). For clustering metrics like Silhouette score, Yake slightly outperforms VN and VR in both the 50 clusters and 100 clusters cases. Similarly, for the Calinski-Harabasz Index (CH) and Davies-Bouldin Index (DBI), Yake achieves slightly better performance than VN, which slightly outperforms VR. Thus overall, in the standard clustering metrics, VN and VR slightly underperform but achieve results comparable to Yake.
%\clearpage
\subsubsection{50 Clusters:}
Table \ref{table2} also shows the context metrics (Mean $\pm$ SD) for the three keyphrase extraction methods. In the case of 50 clusters, VN and Yake achieve nearly identical Unique Word Counts per Cluster, performing better than VR. However, VN and VR have significantly higher Action-Object pairs per cluster (\textgreater 50\% increase), compared to Yake, with VN scoring the highest. As expected, this pattern is exactly the opposite for noun chunks per cluster, where Yake extracts the most, followed by VR and VN. The histograms of Fig. \ref{fig3} (a) help us look at the frequency distribution of these metrics and look beyond the mean values.  For the metrics 'Action-Object pairs' and 'Unique Words,' the distribution of VN's clusters tends to be farther from the origin (more towards the right of the X-axis) than Yake's clusters. But in the case of 'Noun Chunks,' VN's distribution seems to be closest to the origin (more towards the left of the X-axis) than that of Yake. From both these observations, we infer that VN's clusters retrieve more contextual information (more action-object pairs combined with fewer noun chunks). Hence, VN outperforms Yake on nearly all three context metrics. In addition, the peak values(represented by dotted lines) of the distributions  for the three methods roughly correspond with the respective mean values of Table \ref{table2}.

\begin{table}[t]
    \small
    \centering
    \begin{tabular}{|p{0.35\linewidth} | p{0.55\linewidth}| }
    \hline
        \textbf{Category} &  \textbf{Phrase}  \\ \hline
       \multirow{ 3}{*}{Co-branded cards }& Chase amex   \\ \cline{2-2}
        
         & Uber credit card  \\ \cline{2-2}
         & Pass lounge card priority   \\ \cline{2-2}
         & Amazon prime card \\ \hline
      \multirow{ 2}{*} {  Merchant Offers} & Bundle disney credit \\ \cline{2-2}
        
         & Refferal offer bonus   \\ \cline{2-2}
         & Retention offer spend get   \\ \hline
       \multirow{ 2}{*}  {Ad/ Promo Campaigns }& Bce bcp card  \\  \cline{2-2}
& Sub meet spend requirement  \\  \cline{2-2}
 & Air france offer \\ \hline
 \multirow{ 2}{*} {Airport Lounge / } & Lounge delta tsa \\ \cline{2-2}
 & Lax centurion lounge \\  \cline{2-2}
\multirow{ 2}{*}{Concierge Service }& Concierge reservation restaurant  \\  \cline{2-2}
 & Lounge access delta platinum  \\ \hline

    \end{tabular}
    \caption{Summary phrases extracted from different clusters, classified into categories}
    \label{tab:table4}
\end{table}

\subsubsection{100 Clusters:}
In the case of $100$ clusters, similar trends are observed. VN outperforms Yake in terms of unique words per cluster, followed by VR. Like in the case of $50$ clusters, Fig. \ref{fig3} (b) shows that for Action-Object pairs, VN's distribution tends farther from the origin (higher peak value) compared to Yake and so does VR to a smaller extent. Again for noun chunks, VN and VR's distributions are closer to the origin compared to Yake, revealing that Yake's clusters mostly retrieve noun chunks rather than contextual information. Thus, VN again outperforms Yake on all three context metrics. Additionally, since the number of clusters is doubled (from $50$ to $100$), the values of these context metrics are also roughly halved.

In summary, Fig. \ref{fig3} (a) and (b) provide a visual representation of the metric values, their frequencies, and peaks (dotted vertical lines) in the distribution for $50$ and $100$ clusters, respectively. For unique words per cluster, all three methods have similar results, with VN and Yake slightly in the lead. Regarding Action-Object pairs, however, the Yake curve skews to the origin and steeply drops, indicating that its clusters have fewer action-object pairs than those of VN and VR. For Noun chunks, this trend is reversed, with the VN and VR curves skewing towards the origin and Yake having a peak at a greater value. Thus, the combination of higher unique words per cluster, higher action-object pairs, and lower noun chunks establishes that VN outperforms VR, and they both significantly outperform Yake on the context metrics, as well as achieve comparable results on the cluster metrics.

% ********TABLE 3: Qualitative Examples********
\begin{table*}[h]
\small
\centering
\begin{tabular}{p{0.3\linewidth}  p{0.3\linewidth}  p{0.3\linewidth}}
\toprule
\multicolumn{3}{c}{Post}           \\ \midrule
(a) Is the NCFU Navy card eligible for Amex offers?  \textbf{\textless Question\textgreater} 
% Question | Amex Questions
& (b) Marriot Brilliant Losing on-site \$300 credit, replaced with monthly dining \textbf{\textless Observation\textgreater}
% Observation | Amex Deals \& Tips
& (c) Gold Companion Platnium instead of Platinum upgrade offer? Sharing this odd experience with AMEX rep. \textbf{\textless Anecdote\textgreater} 
% Anecdote | Member Inquiry
\\ \midrule

\multicolumn{3}{c}{Thread Summary} \\ \midrule

\textbf{VN:} \textcolor{blue}{offers connect site} ; track card ; add offers faq ; \textcolor{blue}{have offers faq} ; have portal ; \textcolor{blue}{gets small offer} ; be card ; note party cards ;   confirm connect website ; use time
& \textbf{VN:} extend hotels ; \textcolor{blue}{creates effort} ; dining credit ; \textcolor{blue}{maximize card} ; coming marriott bonvoy ; creates lot ; receive dollar credit ; receive dining credit ; made change
& \textbf{VN:} cancelling gold platinum ; \textcolor{blue}{have gold fee} ; \textcolor{blue}{losing card benefits} ; \textcolor{blue}{has fee card} ; enjoy benefits ; get card benefits ; enjoy cards ; including supermarkets ; including x restaurants ; is wrong \\ \midrule

\textbf{VR:} be card ; amex offers ; us network ; american express ; amex site ; nfcu amex ; \textcolor{blue}{good small offer} ; own portal ; \textcolor{blue}{third party cards} ; last year
& \textbf{VR:} \textcolor{blue}{maximize value} ; maximize card ; is change ; brilliant card ; hotel credit ; amex Marriott ; bonvoy card ; dining credit      
& \textbf{VR:} \textcolor{blue}{annual fee card} ; companion card ; get card benefits ; card benefits ; fee card ; companion platinum ; platinum card ; gold card ; is annual gold card \\ \midrule

\textbf{Yake:} \textcolor{red}{American Express} ; American Express OPEN ; American Express Serve ; American Express Consumer ; \textcolor{red}{American Express} Card ; valid \textcolor{red}{American Express} ; AMEX   Offers ; Express Consumer Card; registered \textcolor{red}{American Express} 
& \textbf{Yake:} American Express ; Marriott \textcolor{red}{Bonvoy Brilliant} ; Relevant bit ; \textcolor{red}{Bonvoy Brilliant} card ; Amex Marriott Bonvoy ; \textcolor{red}{Bonvoy Brilliant} ; Amex Marriott card
& \textbf{Yake:} \textcolor{red}{Gold Companion} Platinum ; American Express Gold ; additional Gold card ; Gold card ; \textcolor{red}{Gold Companion} ; Express Gold card ; \textcolor{red}{Companion Platinum} ; additional Gold ; Platinum \\ \bottomrule

\end{tabular}
\caption{Three kinds of posts (tags in \textbf{\textless\textgreater}) and their phrase-based thread summaries, arranged in descending order of importance (according to their scoring functions). VN and VR provide insights from the thread and contextual information (blue). Yake's features, mostly highlight product names and show considerable word repetition (red). The VN, VR and Yake phrases are derived from additional comments (not shown here due to space constraint).}
\label{table3}
\end{table*}

%\vspace{-5mm}

\subsection{Qualitative Examples}
In Table \ref{table3} we present a few qualitative examples of phrase-based summary generation from VN (Verb-Noun Intents), VR (Variant sentiment aspects) and Yake. Here, we show representative posts of three types - Questions, Observations, and Anecdotes. Post (a) is a question about a particular card’s eligibility for offers, where comments directly answer the post. Post (b) is an observation on news about a certain Amex offer, where comments participate in discourse about these changes. Post (c) is an anecdote about a member’s unique experience upgrading their card, and the comments present their opinions on the situation. Yake preferentially extracts noun chunks like product names in its summaries (‘American Express OPEN,’ ‘Bonvoy Brilliant Card,’ etc.) and tends to repeat these keywords. These summaries only provide limited information about the entities involved. 

In contrast, both VN and VR provide more context in their keyphrases. For e.g., in post (a), VN highlights that the particular card is eligible for offers at the ‘CONNECT’ site, and both methods highlight the example of the ‘Small offer.’ Next, In post (b), VN and VR highlight phrases like ‘maximize value’ and ‘create efforts.’ Finally, in post (c), VN highlights a phrase about ‘losing card benefits’ from a comment that clarifies the pros and cons of upgrading. Overall, VR seems to have more repetition than VN in its phrase-based summaries. Both the VN and VR summaries seem to give more useful context information about lengthy threads. This is despite Yake’s higher performance on conventional cluster metrics than VN or VR. Additionally, Table \ref{tab:table4} shows some categorical phrase extraction which represents some important discussions on Reddit related to topics like competitors, co-branded cards, merchant offers and so on. Hence our proposed approach, helps end-users to get a comprehensive view of the aforementioned "external perspective."
% ***************************DISCUSSION***************************

\begin{figure*}[h!]
\centering
    \includegraphics[width = \textwidth]{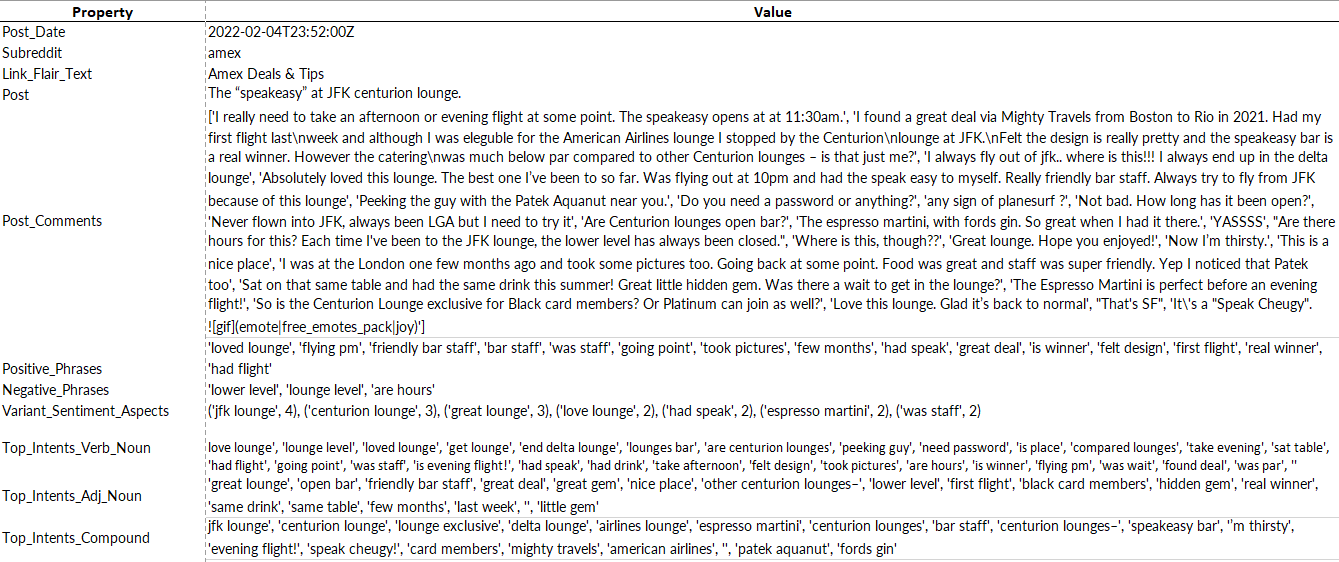}
    \caption{ Screenshot of the web utility portal hosted within American Express}
    \label{fig4}
\end{figure*}

The Quantitative results can largely be attributed to the scoring functions of the respective keyphrase extraction methods used. Yake (the baseline) scores terms using their normalized term frequency \cite{campos2020yake} and prefers terms at the beginning of sentences and those that appear in multiple sentences. Additionally, N-grams are assigned a weighted product of term scores. These facts help to explain the context metrics results for Yake. For instance, the keyphrases generated are necessarily contiguous (N-grams), explaining the prevalence of noun chunks and the lack of action-object pairs. The scoring functions of the VN intents and VR aspects also explain their respective clustering behavior. The scoring function $f()$ for our intent detection method uses average TF-IDF sums of terms to rank intent phrases by the respective grammar rule. However, in the case of the scoring function $g()$ for VR, the ‘aspects’ are drawn from all of the intent phrases (regardless of grammar rule). These phrases are ranked using variance of the sentiment scores of the comments they appear in. Thus, the TF-IDF scores promote the number of unique words per cluster in VN. However, VN and VR both show a higher number of action-object pairs per cluster since they both use the action-object pair generation stage of our pipeline. Figure \ref{fig4} shows the screenshot of web utility portal hosted within Amex, showing the Reddit data (post and comments) along with the different kind of key-phrases extracted for a single Reddit thread.

%\subsection{Further Applications in FinTech }
% {Shreya, please help with this section.}

% \subsection{Limitations}
% In this work, our goal is to address the issue of the lack of labeled data in the FinTech  domain through a completely unsupervised method. However, the lack of ground truth rules out fine-tuning approaches with sequence-to-sequence models and makes evaluation particularly challenging. Additionally, our unsupervised pipeline uses ‘Action-Object’ pairs and depends on pre-defined grammar rules, which may not be generalizable. Finally, the intent phrases (2-4 words) may be too short to capture the complete meaning of certain comments, necessitating a trade-off between brevity and contextual information.

% ***************************CONCLUSION***************************
\section{Conclusion}
\label{sec:conclusion}

This paper discusses problem of unsupervised phrase-based summary generation from social media discussion, particularly in FinTech domain. Social media analysis is an increasingly important tool for financial organizations to understand consumers and identify grievances. However, lack of labeled datasets, expensive annotation procedure, and the restrictive formulation of intent detection as multi-class classification makes this task challenging. To this end, we propose a completely unsupervised summary generation framework using the notion of intent key-phrases (Action-Object pairs) and test it on a dataset of posts and comments we collect from the organization's Reddit community. We use these post-level summaries (intent phrases) to cluster the posts and generate summaries at a group level using agglomerative clustering. As an additional analysis, we also identify aspects (intent phrases) associated with positive and negative sentiments in the comments. We evaluate our framework to Yake, an unsupervised keyword extraction system using clustering metrics. We also propose new metrics to judge the context retrieval in the clusters, and our methods significantly outperform the baseline. Proposed framework has been leveraged as a web utility portal hosted within Amex and successfully being used for research purposes. Future studies may investigate different embedding techniques to represent the intent phrases and seq2seq keyphrase generation models applied to even bigger social media datasets. 
%Our results are comparable to the baseline for typical clustering metrics. 

%Finally, we discuss the implications of these unsupervised methods in the FinTech industry and their potential applications. 
\vspace{10mm}

% Entries for the entire Anthology, followed by custom entries
%\bibliography{anthology,custom}
\bibliographystyle{ACM-Reference-Format}
%%% -*-BibTeX-*-
%%% Do NOT edit. File created by BibTeX with style
%%% ACM-Reference-Format-Journals [18-Jan-2012].

% \appendix

% \section{Example Appendix}
% \label{sec:appendix}

% This is an appendix.

\end{document}